\documentclass[runningheads]{llncs}
\usepackage[camera-ready,year=2024,ID=8535]{eccv}
\usepackage{eccvabbrv}
\usepackage{graphicx}
\usepackage{booktabs}
\usepackage{multirow}
\usepackage{tabularx}
\usepackage[accsupp]{axessibility}

\usepackage[pagebackref,breaklinks,colorlinks,citecolor=eccvblue]{hyperref}

\usepackage{orcidlink}

\begin{document}

\title{PMT: Progressive Mean Teacher via Exploring Temporal Consistency for Semi-Supervised Medical Image Segmentation} 

\titlerunning{Progressive Mean Teacher}

\author{Ning Gao\orcidlink{0009-0007-1431-2629} \and
Sanping Zhou$^{*}$\orcidlink{0000-0003-4100-2304} \and
Le Wang\orcidlink{0000-0001-7073-1050} \and
Nanning Zheng\orcidlink{0000-0003-1608-8257}}

\authorrunning{N.~Gao et al.}

\institute{
National Key Laboratory of Human-Machine Hybrid Augmented Intelligence,
National Engineering Research Center for Visual Information and Applications,
Institute of Artificial Intelligence and Robotics, Xi’an Jiaotong University
}

\maketitle

\renewcommand{\thefootnote}{\fnsymbol{footnote}}
\footnotetext[1]{Corresponding author.}

\begin{abstract}
  Semi-supervised learning has emerged as a widely adopted technique in the field of medical image segmentation. The existing works either focuses on the construction of consistency constraints or the generation of pseudo labels to provide high-quality supervisory signals, whose main challenge mainly comes from how to keep the continuous improvement of model capabilities. In this paper, we propose a simple yet effective semi-supervised learning framework, termed \textbf{P}rogressive \textbf{M}ean \textbf{T}eachers~(PMT), for medical image segmentation, whose goal is to generate high-fidelity pseudo labels by learning robust and diverse features in the training process. 
  Specifically, our PMT employs a standard mean teacher to penalize the consistency of the current state and utilizes two sets of MT architectures for co-training. The two sets of MT architectures are individually updated for prolonged periods to maintain stable model diversity established through performance gaps generated by iteration differences. Additionally, a difference-driven alignment regularizer is employed to expedite the alignment of lagging models with the representation capabilities of leading models. Furthermore, a simple yet effective pseudo-label filtering algorithm is employed for facile evaluation of models and selection of high-fidelity pseudo-labels outputted when models are operating at high performance for co-training purposes.
  Experimental results on two datasets with different modalities, i.e., CT and MRI, demonstrate that our method outperforms the state-of-the-art medical image segmentation approaches across various dimensions. 
  The code is available at \url{https://github.com/Axi404/PMT}.
\end{abstract}

\section{Introduction}
\label{sec:intro}

\indent\indent Semi-supervised learning is an important field in deep learning, which offers an effective way to tackle problems with limited labeled data~\cite{zhou2021multinetwork,zheng2022simmatch,rizve2022towards,xia2023learning,xin2019semi}. With the continuous emergence of large-scale data, semi-supervised learning has become a hot topic in both machine learning and computer vision communities. It has gained particular attention in the medical image segmentation~\cite{wang2020meta}, due to the need for expert annotation in determining the accurate boundaries of targets. As a result, the labeled data is very lacking in medical image segmentation, which makes the semi-supervised learning very popular in this domain.

\begin{figure}[t]
  \centering
  \includegraphics[width=0.65\linewidth]{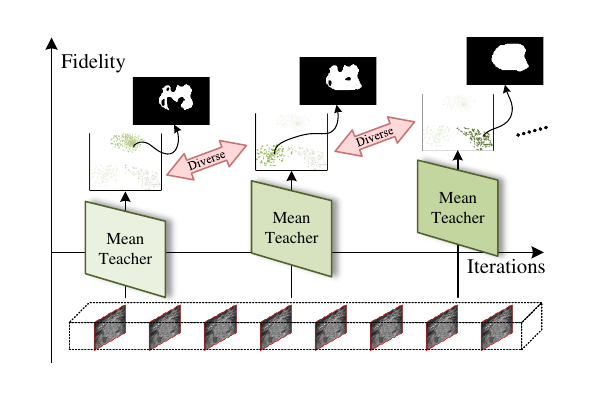}
  \caption{Motivation of our PMT. In particular, the standard MT helps to learn robust features by keeping the consistency between teacher and student networks at the current iteration, while our PMT further helps to learn diverse features by maintaining the difference between student networks at different iterations. As a result, more and more high-fidelity pseudo labels will be generated for semi-supervised medical image segmentation.}
  \label{fig:motivation}
\end{figure}

Most of the semi-supervised learning methods involve seeking a guidance mechanism using limited annotated data, so as to maximize the usage of unlabeled data to train the models on different downstream tasks. In general, two technologies, \emph{i.e.}, consistency regularization~\cite{laine2016temporal,lee2013pseudo} and pseudo label generation~\cite{lee2013pseudo,tarvainen2017mean}, are commonly used in semi-supervised medical segmentation. The former ones aim at enhancing the representation capability of networks by exploring the consistency based on different prior assumptions~\cite{chen2021exploring}, while the latter ones focus on improving the performance on downstream tasks by generating the pseudo labels based on current model capability. For example, the well-known Mean Teacher~(MT)~\cite{tarvainen2017mean} utilizes an ingenious Exponential Moving Average~(EMA) that is equivalent to data augmentation and enforces consistency regularization between the outputs of teacher and student models. Besides, the  representative Noisy Student~\cite{xie2020self} designs a teacher model to generate pseudo labels, so as to train a larger student model.

Even though the above methods have achieved significant improvements in semi-supervised medical image segmentation, we still argue that they are weak in exploring high-quality supervisory signals to consistently enhance the model's capability. To address this issue, an iterative unified optimization framework has been introduced to semi-supervised medical image segmentation~\cite{wang2023mcf,wu2022cross,wang2023dhc}, in which the consistency regularization strategy is taken to enhance the model's representation capability and the pseudo label generation algorithm is applied to generate high-fidelity pseudo labels. For example, the recent MCF~\cite{wang2023mcf} takes VNet and 3D ResNet for representation learning, and heterogeneous networks for dynamic pseudo label generation. The challenges to these methods lies on how to continuously generate diverse pseudo labels in the forward propagation, and enhance model's capability in the backward propagation. 

In this paper, we design a novel semi-supervised learning framework, termed Progressive Mean Teachers~(PMT), for medical image segmentation, whose main idea focuses on how to obtain a diverse set of accurate pseudo labels. Inspired by the positive relationship between iteration and performance, we try to maintain the diversity of networks by exploring their states at different training epochs. Specifically, the standard MT is first taken as basic architecture to learn the parameters of network at each iteration, which can enhance the network's representation capability by using the EMA data augmentation. Our PMT further explores network diversity during training, alternating between two homogeneous MT architectures trained on the same dataset, a process we term progressive design. These models exhibit significant iteration leads due to alternating continuous individual updates, establishing performance gaps between networks at different epochs, as shown in \textbf{Fig.~\ref{fig:motivation}}. Consequently, the student network can acquire robust yet diverse features for medical image segmentation.
Then, the Discrepancy Driven Alignment~(DDA) regularizer is further designed to examine disparities between predictions obtained by the student network at different training epochs, facilitating rapid alignment to high-fidelity generated images. Finally, we design a simple Pseudo Label Filtering (PLF) algorithm to refine the basic interaction process, enabling the retention of high-fidelity pseudo-labels for training by comparing student network performance across different training epochs. As a result, more and more high-fidelity pseudo labels can be fed to train the other student networks, which  will in turn to generate more accurate predictions for pseudo label generation.

In summary, the main contributions of this work are as follows:~(1) We design a novel Progressive Mean Teacher framework for semi-supervised medical image segmentation. (2) We design a novel Discrepancy Driven Alignment regularizer
to rapidly align the representational capacity gap between lagging and leading networks.
(3) We design a simple yet effective Pseudo Label Filtering algorithm to select high-fidelity pseudo labels. Extensive experiments in both Left Atrial~\cite{xiong2021global} and Pancreas-NIH~\cite{roth2015deeporgan} datasets show that our PMT can achieve the state-of-the-art results in semi-supervised medical image segmentation.

\section{Related Work}
\label{sec:RelatedWork}

\subsection{Consistency Regularization} 

\indent\indent Consistency regularization is often employed in semi-supervised learning, so as to enhance the stable representational capacity of model. The behind idea is to preserve the invariance of predictions made by the same model when facing perturbations applied in different regions, by constraining the consistency between output results under different perturbations. For example, $\Pi$ model~\cite{laine2016temporal} introduces image-level regularization by adding perturbations to images. The well-known MT~\cite{tarvainen2017mean} introduces parameter-level regularization by using EMA to constrain the outputs between teacher and student models. Thanks to its simplicity and effectiveness, more and more works are paying attention to improve the generalization ability by formulating different consistency regularizers. For example, SASSnet~\cite{li2020shape} focuses on the regularity of geometric shapes for target object classes within consistency regularization. Besides, CPCL~\cite{xu2022all} establishes regularization between supervised and unsupervised training within a cyclic framework. Furthermore, some later works have made progress in regularization at different task and model levels. For example, DTC~\cite{luo2021semi} introduces task-level regularization, presenting a novel dual-task consistency semi-supervised framework. Besides, MCF~\cite{wang2023mcf} introduces model-level regularization by using heterogeneous models to constrain output consistency. Compared to previous work, our PMT seeks the cross-temporal regularization between different  training periods, which can help learn diverse yet robust features for subsequent pseudo label generation.

\subsection{Pseudo Label Generation}

\indent\indent Pseudo label generation is often employed in semi-supervised learning, so as to enhance the discriminative ability of model. The behind idea is to train a prior model with the labeled data and then apply it to generate the pseudo labels for unlabeled data. It is widely accepted to categorize pseudo-label generation into two methods~\cite{jiao2022learning}, direct generation focuses on selecting pseudo labels with higher confidence, while indirect generation explores methods to generate high-fidelity pseudo labels. For direct generation, work~\cite{lee2013pseudo} uses a fixed threshold to choose high-confidence pseudo labels. SsaNet~\cite{wang2022ssa} employs a trust module to reevaluate pseudo labels. UA-MT~\cite{yu2019uncertainty} introduces uncertainty estimation to filter out unreliable pseudo labels. Co-BioNet~\cite{peiris2023uncertainty} introduces a feedback network to measure the uncertainty and choose high-confidence predictions of different models. For indirect generation, Tri-Net~\cite{dong2018tri} is proposed to use two subnetworks to generate pseudo labels for a third subnetwork. Work~\cite{achanta2012slic} improves pseudo label generation using Simple Linear Iterative Clustering~(SLIC) algorithm. MCF~\cite{wang2023mcf} dynamically generates pseudo labels using a heterogeneous network and effectively addresses cognitive bias, while DeSCO~\cite{cai2023orthogonal} focuses on the spatial correlation of medical images and generates pseudo labels using orthogonal slices. Compared to previous work, our PMT emphasizes the enhancement of pseudo label quality in the temporal domain and generates diverse and robust pseudo labels across different training periods, significantly improving performance.

\subsection{Multi-Model Framework}

\indent\indent The multi-model framework is often employed in semi-supervised learning, so as to enhance the model representation by acquiring multiple views or diversity. Its development aligns closely with consistency regularization. Here, we focus on the evolution of model structures. For example, MT~\cite{tarvainen2017mean} introduces a Teacher model with significantly improved representation capabilities at a lower cost using EMA, establishing the MT architecture. Besides, CPC~\cite{ke2020guided} utilizes the confidence vector of multi-model outputs as pseudo-labels for co-training, while CPS~\cite{chen2021semi} employs one-hot labels, both considered as starting points for co-training. In recent years, efforts have been made to combine these approaches. For example, UCMT~\cite{shen2023co} employs two student models for co-training, simultaneously updating the same Mean Teacher using EMA, while Dual Teacher~\cite{na2024switching} alternates updates between two Teacher models using EMA with a single student model, thereby fostering diversity among Teacher models. Compared to previous work, our PMT utilizes two sets of Mean Teachers in a progressive training framework, and employ two student models to update teacher models independently. This enables our model to rapidly establish robust performance disparities across iteration gap and maintain stable diversity among models.

\section{Method}

\label{sec:Method}

\subsection{The Overall Process of PMT}

\begin{figure*}[t]
  \centering
  \includegraphics[width=1.0\linewidth]{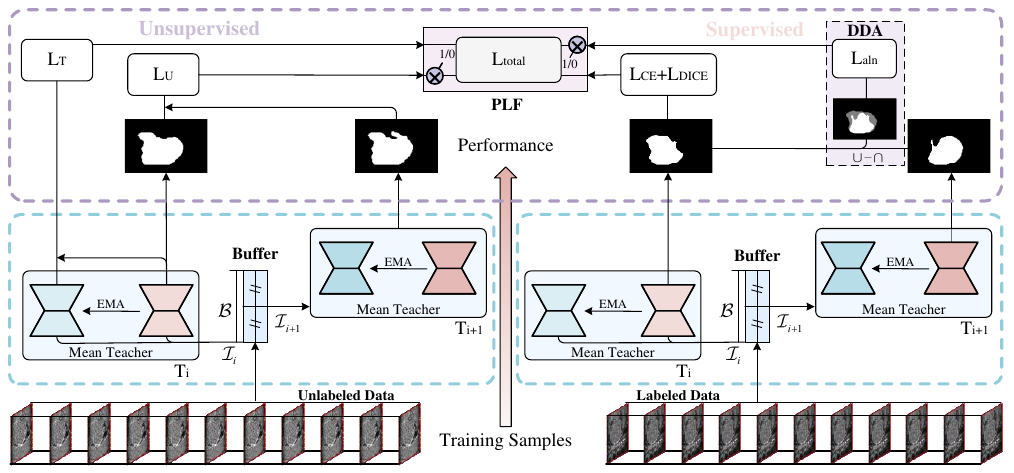}
  \caption{An overview of PMT. We employed a progressive design and utilized the architecture of MT. The PMT framework maintains a data buffer of length $\mathcal B$ for cross-temporal training. The total loss function $\mathrm L_\mathrm{total}$ for each network includes supervised losses $\mathrm L_\mathrm{CE}, \mathrm L_\mathrm{DICE}, \mathrm L_\mathrm{aln}$, and unsupervised loss $\mathrm L_\mathrm{U}$,  $\mathrm L_\mathrm{T}$.}
  \label{fig:pt}
\end{figure*}

\indent\indent The training process of PMT is illustrated in \textbf{Fig.~\ref{fig:pt}}. As previously mentioned, PMT can consist of multiple networks. For ease of explanation, the number of networks is assumed to be two by default in our work. In our approach, these networks share the same structure and can be denoted as $f_i(\cdot)\in\mathcal{F}$, where $\mathcal{F}$ stands for the function space within which these networks reside. In a semi-supervised process, the training data includes a small number of labeled data denoted as $\mathbf D_L = \{(x_i^L, y_i^L)\}_{i=1}^N$, and a large amount of unlabeled data denoted as $\mathbf D_U = \{(x_i^U)\}_{i=N+1}^{N+M}$, where $N \ll M$, $x_i \in \mathbb{R}^{H \times W \times D}$ represents medical volumes, and $y_i \in \{0,1\}^{H \times W \times D}$ represents ground truth labels. Batches of input data $\mathbf X$ consist of an equal proportion of labeled data $(\mathbf X^L, \mathbf Y^L)$ and unlabeled data $\mathbf X^U$. These volumes are fed into $f_i(\cdot)$ and $f_{i+1}(\cdot)$:

\begin{equation}
  \hat{\mathbf Y}_{i} = f_{i} (\mathbf X), \quad\hat{\mathbf Y}_{i+1} = f_{i+1} (\mathbf X).
  \label{eq:subnet output}
\end{equation}
The output consists of volume predictions for both labeled and unlabeled data: $\hat{\mathbf Y} = \hat{\mathbf Y}^L \cup \hat{\mathbf Y}^U$. For the sake of simplicity, the network index subscripts are omitted here. 

We have ingeniously reintroduced MT architecture into the framework of semi-supervised learning. The introduction serves the purpose that, while providing high-fidelity pseudo labels to the model, the model's architecture needs to support the stable improvement of its representational capacity, aiming for better performance, while maintaining stable diversity. 
The teacher network is structurally identical to the student network but does not actively participate in the training process. Instead, it updates all of its parameters through EMA and applies consistency regularization to the student model through its output. The parameter is updated as follows:

\begin{equation}
    \theta_{\text{teacher}} = \alpha \theta_{\text{teacher}} + (1-\alpha) \theta_{\text{student}},
    \label{eq:ema update}
\end{equation}
where $\theta_{\text{teacher}}$ represents the parameters of teacher network, $\theta_{\text{student}}$ represents the parameters of student network, and $\alpha$ is the EMA decay rate.

In practice, we propose a cross-temporal training approach, which involves introducing a phase shift among different models during iterations and training the most lagging model in terms of iterations. To ensure that each model can train across temporal, the number of iterations for the model recorded in the sequence of iterations is denoted as $\mathcal {I}_i$. Our model maintains a static iteration gap through the data buffer length $\mathcal B$:

\begin{equation}
    \forall i \in [1,n-1],\text{Lar}(\mathcal I,i) - \text{Lar}(\mathcal I,i+1) = \frac{\mathcal{B}}{(n-1)},
\end{equation}
where, $\text{Lar}(\mathcal I,i)$ represents the $i$-th largest number in $\mathcal I$.

During each training iteration, the model with the least advanced iteration progress, referred to as the Current Progressive Model (CPM), sequentially utilizes data from the buffer for training until it has advanced ahead of the model with the most advanced iteration progress $\mathcal{B}/n$ times. Throughout the training process, the PLF screens out pseudo-labels generated by models whose performance is inferior to that of the CPM by measuring the performance gap between the CPM and other models, and learns from the remaining pseudo-labels. Simultaneously, the model computes the DDA with respect to a given input $\mathbf{X}$ and, along with all other models, examines regions of inconsistent predictions among different networks. When the performance of the CPM lags behind that of other models, the DDA corrects the CPM, enabling it to quickly align its performance with models that outperform it. It is worth noting that if the CPM is the best-performing model for that iteration, it will not guide other models in reverse. It receives guidance during the training process and only guides others once they have completed training.

\begin{figure}[t]
  \centering
  \includegraphics[width=0.6\linewidth]{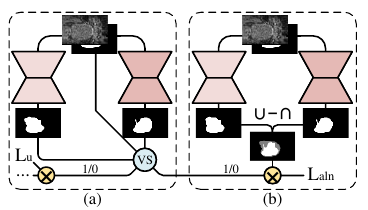}
  \caption{Illustration of PLF and DDA. PLF is utilized for comparing the representational capacity of models, while DDA aligns model outputs by examining differences.}
  \label{fig:PLF}
\end{figure}

\subsection{Pseudo Label Filtering}

\indent\indent To guide the learning of the current progressive model effectively, we introduce PLF, as shown in \textbf{Fig.~\ref{fig:PLF}}~(a). PLF involves a naive approximation where we consider the representational capacity of models during the supervised learning phase as a criterion for judging the representational capacity during the unsupervised learning phase to some extent. During the supervised learning phase, PLF entails inference by all models, and through combining labels, calculates the representational capacity of each model in the current iteration. The metrics for evaluating representational capacity vary with tasks; in this task, the Dice loss is regarded as a statistical measure of model performance. We utilize the representational capacity of the CPM as a threshold and filter out pseudo-labels generated by models with representational capacities below that of the CPM. 
The pseudo label's passage through the PLF is recorded as $\mathcal{P}(f_i,f_{i+1})$, where a value of 1 denotes passage and 0 denotes exclusion.
This aids in the model's better learning of pseudo-labels with high fidelity. Furthermore, to ensure that the pseudo-labels are accurate and well-defined, we apply a sharpening function to high-fidelity pseudo-labels:

\begin{equation}
\mathbf Y^U = \frac{\mathbf {P_s}^{1/T}}{\mathbf {P_s}^{1/T} + (1-\mathbf {P_s})^{1/T}},
\label{eq:sharpen}
\end{equation}
where $\mathbf {P_s}$ represents the output of the good student, and $T$ is a hyperparameter.

The resulting loss function $\mathrm L_\mathrm U$ is computed as MSE between the pseudo labels $\mathbf Y^U$ and the model's predictions of CPM $\hat{\mathbf Y}^U_c$:

\begin{equation}
\mathrm L_\mathrm U = \operatorname{MSE}(\hat{\mathbf Y}_c^U, \mathbf Y^U).
\end{equation}

\subsection{Discrepancy Driven Alignment}

\indent\indent We propose DDA to compare the outputs of multiple models and achieve rapid alignment between models, as shown in \textbf{Fig.~\ref{fig:PLF}}~(b). During the alternating process of progressive Learning, the performance of two groups of models alternately improves, providing stable pseudo-labels to each other. However, concerns often arise due to the disappearance of the performance gap between models caused by random factors. While the expectation is for alternating performance leadership between models, in practice, it may turn into one model consistently leading. Therefore, a fast and high-confidence alignment module is needed. In traditional consistency regularization, contrastive algorithms enforce consistency among the outputs of different models. However, this approach often raises concerns because we are uncertain whether this consistency regularization guides the models in the right direction. DDA focuses on the different parts of predictions among models, which often reflect the differences in representational capabilities between models. We aim to enhance the consistency regularization specifically on these parts to achieve alignment of representational capabilities between models. It is worth noting that we expect the models aligned by DDA to have representation capabilities surpassing CPM. Therefore, DDA will only take effect when PLF allows other models to provide pseudo-labels. The mask of different parts of predictions among different models can be obtained by taking the difference between the union and intersection of the binary outputs of softmax outputs $\hat{\mathbf Y}^{L}$ of multiple networks:

\begin{equation}
    \mathcal M_{\mathrm{diff}} = \bigcup_{i=1}^{n} \operatorname{BINA}(\hat{\mathbf{Y}}_{i}^{L}) - \bigcap_{i=1}^{N} \operatorname{BINA}(\hat{\mathbf{Y}}_{i}^{L}).
    \label{eq:mdiff}
\end{equation}

Subsequently, MSE of the model's predictions is computedand denoted as $\mathcal M_\mathrm{dist}$:

\begin{equation}
    \mathcal{M}_\mathrm{dist} = \operatorname{MSE}\left(\hat{\mathbf Y}_{i}^{L}, {\mathbf Y}^{L}\right).
    \label{eq:msedist}
\end{equation}

Next, the total count of elements in $\mathcal M_{\mathrm {diff}}$ is calculated, and $\mathcal M_{\mathrm {diff}}$ is used to mask $\mathcal M_{\mathrm {dist}}$. The ratio of the sum of the masked $\mathcal M_{\mathrm {dist}}$ to the total count of elements in $\mathcal M_{\mathrm {diff}}$ is denoted as $\mathrm L_\mathrm{aln}$, which characterizes the gap to the ground truth within the prediction differences between two models:

\begin{equation}
    \mathrm L_\mathrm{aln} = \frac{\sum \left(\mathcal M_{\mathrm{diff}} \cdot \mathcal{M}_\mathrm{dist}\right)}{\sum \left(\mathcal M_{\mathrm{diff}}\right)}.
    \label{eq:lret}
\end{equation}

\subsection{Loss Function}

\indent\indent In general, during an iteration cycle, CPM conducts the process of backward propagation concerning its loss function, which is defined as follow:
\begin{equation}
\mathrm L_\mathrm{total} = \mathrm L_\mathrm s + \lambda_1  \mathcal{P}(f_i,f_{i+1})\mathrm L_\mathrm U + \lambda_2 \mathrm L_\mathrm T,
\end{equation}
where $\lambda_1$ and $\lambda_2$ increase as the number of iterations grows, up to a point where they stop growing after a fixed iterations. 

We employed two independent Gaussian warm-up function to control the loss function weights, $\lambda_1$ and $\lambda_2$, using different parameters:

\begin{equation}
\begin{aligned}
&\lambda_1(t) = \begin{cases} 
\hat\lambda_{1} \cdot e^{-5\left(1-\frac{2t}{t_{\max}}\right)^2}, &  t < \frac{t_{\max}}{2} \\
\hat\lambda_{1}, &  t \geq \frac{t_{\max}}{2}
\end{cases}\\
&\lambda_2(t) = \begin{cases} 
\hat\lambda_{2} \cdot e^{-5\left(1-\frac{2t}{t_{\max}}\right)^2}, &  t < \frac{t_{\max}}{2} \\
\hat\lambda_{2}, &  t \geq \frac{t_{\max}}{2}
\end{cases}
\end{aligned}\end{equation}
where $t$ represents the current iteration number, and $t_{\max}$ represents the total number of training iterations. The hyperparameter $\hat\lambda_{1}$ and $\hat\lambda_{2}$  were empirically set to 20.0 and 10.0.

Loss function $\mathrm L_\mathrm s$ represents the loss during supervised learning and consists of the following components:

\begin{equation}
\mathrm L_\mathrm s = \operatorname{CE} + \operatorname{DICE} + \beta\mathcal{P}(f_i,f_{i+1}){\mathrm L_\mathrm{aln}}(\mathcal M_{\text{diff}}, \mathcal{M}_{dist}),
\end{equation}
where $\beta$, a constant set to 0.5, is used to balance alignment loss and other losses.

Loss function $\mathrm L_\mathrm T$ is the consistency loss generated by the MT, which is derived by the teacher through the sharpen function to convert the model outputs into pseudo-labels, and subsequently generated by computing the MSE with the student model:

\begin{equation}
    \mathrm L_\mathrm T = \mathrm{MSE}\left(f_i\left(\mathbf X^U\right),\frac{\mathbf {P_t}^{1/T}}{\mathbf {P_t}^{1/T} + (1-\mathbf {P_t})^{1/T}}\right).
\end{equation}

\label{sec:Experiments}
\section{Experiments}

\subsection{Implementation Details}

\indent\indent We selected the VNet~\cite{milletari2016v} model as a baseline network, which performs well in conditions with limited data and is essentially a 3D convolutional version of UNet. During inference, we use the average of the outputs from two networks as the final prediction. Specifically, the SGD optimizer was used to update the network parameters with weight decay of $0.0001$ and a momentum of $0.9$. The initial learning rate was set to $0.01$, divided by $10$ every $2500$ iterations, for a total of $6000$ iterations. 

Following the practice in comparative literature~\cite{yu2019uncertainty,li2020shape,luo2021semi,wang2023mcf}, our methods are trained for a fixed number of 6,000 iterations to obtain the final model. Additionally, our models all use a batch size of 4, with a labeled data quantity of 2. We tested the performance of the models, and all experiments were conducted on $\text{NVIDIA}^\text{\textregistered}$ GeForce A40 48GB running Ubuntu 20.04 and PyTorch 1.11.0.

\subsection{Datasets and Metrics}

\indent\indent In the experiment, we selected two datasets with different modalities and utilized four distinct metrics to assess the performance of the model. For each dataset, 80\% of the data was used as the training set and 20\% as the test set. The proportion of supervised data was determined based on the training set.

\textbf{LA Dataset.} It~\cite{xiong2021global} includes 100 3D gadolinium-enhanced MR imaging volumes of left atrial with an isotropic resolution of $0.625 \times 0.625 \times 0.625{mm}^3$ and the corresponding ground truth labels. For pre-processing, we first normalize all volumes to zero mean and unit variance, then crop each 3D MRI volume with enlarged margins according to the targets. During training, the training volumes are randomly cropped to $112 \times 112 \times 80$ as the model input. During inference, a sliding window of the same size is used to obtain segmentation results with a stride of $18 \times 18 \times 4$. 

\textbf{Pancreas-NIH Dataset.} It~\cite{roth2015deeporgan} provides 82 contrast-enhanced abdominal 3D CT volumes of pancreas with manual annotation. The size of each CT volume is $512 \times 512 \times \mathrm D$, where $\mathrm D \in \left[181, 466\right]$. In pre-processing, we use the soft tissue CT window of $\left[ - 120, 240\right]$ HU, and we crop the CT scans centering at the pancreas region, and enlarge margins with 25 voxels. The training volumes are randomly cropped to $96 \times 96 \times 96$ as the model input. During inference, a sliding window of the same size is used to obtain segmentation results with a stride of $16 \times 16 \times 16$.

\textbf{Metrics.} Following \cite{wang2023mcf,yu2019uncertainty,luo2021semi,xu2022all,bai2023bidirectional,li2020shape}, we use four metrics to evaluate model performance, including regional sensitive metrics: Dice similarity coefficient (Dice)~\cite{yu2019uncertainty}, Jaccard similarity coefficient (Jaccard)~\cite{luo2021semi}, and edge sensitive metrics: 95\% Hausdorff Distance (95HD)~\cite{xu2022all} and Average Surface Distance (ASD)~\cite{bai2023bidirectional}.

\subsection{Ablation Study}

\indent\indent For simplicity, we conducted ablation experiments on LA dataset to evaluate our design choices for each component. For a reliable assessment, all other parts were kept consistent except for the component under investigation.

\textbf{Analysis of PMT Framework.} The PMT architecture under progressive design is at the core of our work, wherein progressive design ensures the continuous generation of diverse pseudo-labels during the forward propagation process. Within the PMT architecture, PLF and DDA respectively denote the pseudo-label filtering and model alignment methods proposed in our paper, while MT ensures stable enhancement of model capability. We employ a non-progressive design co-training framework consisting of two VNet components as the baseline model. When adding PLF and DDA methods to the baseline model, the progressive design is simultaneously incorporated into the baseline model. Finally, we evaluate the impact of the MT architecture on model representation capability, with results presented in \textbf{Table~\ref{tab:MP}}. Overall, compared to the baseline model, which already possesses good representation capability, the PMT framework greatly enhances the model's representation capability.

\begin{table*}[h]
  \centering
  \caption{Ablation results about \textbf{PMT framework} on LA dataset}
  \setlength{\tabcolsep}{0pt}
  \begin{tabularx}{0.92\textwidth}{@{}>{\centering\arraybackslash}p{0.09\textwidth}>{\centering\arraybackslash}p{0.09\textwidth}>{\centering\arraybackslash}p{0.09\textwidth}>{\centering\arraybackslash}p{0.13\textwidth}>{\centering\arraybackslash}p{0.13\textwidth}>{\centering\arraybackslash}p{0.13\textwidth}>{\centering\arraybackslash}p{0.13\textwidth}>{\centering\arraybackslash}p{0.13\textwidth}@{}}
    \toprule
    \multicolumn{3}{c}{Method} &\multirow{2}{*}{Labeled} &\multicolumn{4}{c}{Metrics} \\
    \cmidrule(lr){1-3} \cmidrule(lr){5-8}
    PLF& DDA & MT& &Dice$\uparrow$ & Jaccard$\uparrow$ & 95HD$\downarrow$ & ASD$\downarrow$ \\
    \midrule
    - & -&- &8(10\%) &88.28 &79.31 &7.59 &2.34 \\
    \checkmark &- &- &8(10\%) &89.90 &81.73 &6.14 &1.72 \\
   - & \checkmark &- &8(10\%) &{90.43} &{82.60} &5.98 &1.66 \\
    \checkmark & \checkmark &- &8(10\%) &90.43 &82.60 &\textbf{5.49} &\textbf{1.49} \\
    \checkmark &\checkmark &\checkmark &8(10\%) &\textbf{90.81} &\textbf{83.23} &{5.61} &{1.50} \\
    \bottomrule
  \end{tabularx}
  \label{tab:MP}
\end{table*}

\textbf{Analysis of Regularization Strength.} The hyperparameters $\hat\lambda_{1}$ and $\hat\lambda_{2}$ respectively characterize the regularization strength of the progressive methods and MT architecture on the model. We use a tenfold scale and scale $\hat\lambda_{1}$ and $\hat\lambda_{2}$ based on the hyperparameter specifications we use, as shown in \textbf{Table~\ref{tab:lambda}}. The results indicate that variations in the two parameters within a certain range do not significantly affect the model's performance, demonstrating a certain level of robustness of our model to parameter variations. Within a certain range, on the LA dataset with 10\% labeled data, our model achieves the best performance when $\hat\lambda_{1}$ and $\hat\lambda_{2}$ are 20.0 and 10.0 respectively.

\begin{table*}[h]
  \centering
  \caption{Ablation results about \textbf{regularization strength} on LA dataset}
  \setlength{\tabcolsep}{0pt}
  \begin{tabularx}{0.9\textwidth}{@{}>{\centering\arraybackslash}p{0.09\textwidth}>{\centering\arraybackslash}p{0.09\textwidth}>{\centering\arraybackslash}p{0.2\textwidth}>{\centering\arraybackslash}p{0.13\textwidth}>{\centering\arraybackslash}p{0.13\textwidth}>{\centering\arraybackslash}p{0.13\textwidth}>{\centering\arraybackslash}p{0.13\textwidth}@{}}
    \toprule 
    \multicolumn{2}{c}{$\lambda$} & \multirow{2}{*}{Labeled} & \multicolumn{4}{c}{Metrics} \\
    \cmidrule(lr){1-2}  \cmidrule(lr){4-7}
    $\lambda_{1max}$ & $\lambda_{2max}$ &  & Dice$\uparrow$ & Jaccard$\uparrow$ & 95HD$\downarrow$ & ASD$\downarrow$ \\
    \midrule
    2.0 & 1.0 &8(10\%) &89.66 &81.33 &6.35 &1.84 \\
    2.0 & 10.0 &8(10\%) &{90.47} &{82.66} &\textbf{5.58} &1.64 \\
    2.0 & 100.0 &8(10\%) &87.74 &78.28 &9.33 &2.84 \\
    20.0 & 1.0 &8(10\%) &90.67 &82.98 &6.31 &{1.77} \\
    20.0 & 10.0 &8(10\%) &\textbf{90.81} &\textbf{83.23} &{5.61} &\textbf{1.50} \\
    20.0 & 100.0 &8(10\%) &88.00 &78.73 &9.05 &2.78 \\
    200.0 & 1.0 &8(10\%) &89.08 &80.42 &11.40 &{3.07} \\
    200.0 & 10.0 &8(10\%) &89.75 &81.47 &7.08 &2.22 \\
    200.0 & 100.0 &8(10\%) &83.80 &72.55 &16.44 &5.11 \\
    \bottomrule
  \end{tabularx}
  \label{tab:lambda}
\end{table*}

\subsection{Comparison with Other Methods}

\indent\indent We compared our approach with previous state-of-the-art methods on LA dataset and Pancreas-NIH dataset.

We chose VNet as baseline models for comparison. For the selected alternative models, we opted for UA-MT~\cite{yu2019uncertainty} with uncertainty estimation, SASSNet~\cite{li2020shape} focusing on the regularity of geometric shapes, DTC~\cite{luo2021semi} with task-level regularization, BCP~\cite{bai2023bidirectional} using bidirectional CutMix~\cite{yun2019cutmix}, and MCF~\cite{wang2023mcf} with model-level regularization, with BCP and MCF being state-of-the-art results. Noting that, for BCP, we follow its parameter settings for pre-training 2,000 times and self-training 15,000 times.

{\textbf{Comparison on LA Dataset.}  We conducted a cross-model comparison on the classic LA dataset. We tested with 5\% and 10\% of labeled data. Results of the experiments are presented in \textbf{Table~\ref{tab:result_LA_5}}. To provide a more intuitive demonstration of the performance of various models on the LA dataset, we have selected some representative results for visualization, as illustrated in \textbf{Fig.~\ref{fig:la2}}. Areas with inaccurate segmentation have been annotated accordingly.

\begin{table*}[h]
  \centering
  \setlength{\tabcolsep}{0pt}
  \caption{Comparison results on LA dataset with 5\% and 10\% labeled data}
  \begin{tabularx}{0.92\textwidth}{@{}>{\centering\arraybackslash}p{0.25\textwidth}>{\centering\arraybackslash}p{0.15\textwidth}>{\centering\arraybackslash}p{0.13\textwidth}>{\centering\arraybackslash}p{0.13\textwidth}>{\centering\arraybackslash}p{0.13\textwidth}>{\centering\arraybackslash}p{0.13\textwidth}@{}}
    \toprule
    \multirow{2}{*}{Method} & \multirow{2}{*}{Labeled} & \multicolumn{4}{c}{Metrics} \\
     \cmidrule(lr){3-6}
    &  & Dice$\uparrow$ & Jaccard$\uparrow$ & 95HD$\downarrow$ & ASD$\downarrow$ \\
    \midrule
    VNet &4(5\%) &52.55 &39.60 &47.05 &9.87  \\
    UA-MT &4(5\%) &82.26 &70.98 &13.71 &3.82 \\
    SASSNet &4(5\%) &81.60 &69.63 &16.16 &3.58 \\
    DTC &4(5\%) &81.25 &69.33 &14.90 &3.99 \\
    $\text{MCF}^{\text{SOTA}}$ &4(5\%) &- &- &- &- \\
    $\text{BCP}^{\text{SOTA}}$ &4(5\%) &88.02 &78.72 &7.90 &2.15 \\
    PMT(Ours) &4(5\%) &\textbf{89.47} &\textbf{81.04} &\textbf{6.45} &\textbf{1.86} \\
    \midrule
    VNet &8(10\%) &82.74 &71.72 &13.35 &3.26 \\
    UA-MT &8(10\%) &86.28 &76.11 &18.71 &4.63 \\
    SASSNet &8(10\%) &85.22 &75.09 &11.18 &2.89 \\
    DTC &8(10\%) &87.51 &78.17 &8.23 &2.36 \\
    $\text{MCF}^{\text{SOTA}}$ &8(10\%) &88.71 &80.41 &6.32 &1.90 \\
    $\text{BCP}^{\text{SOTA}}$ &8(10\%) &89.62 &81.31 &6.81 &{1.76} \\
    PMT(Ours) &8(10\%) &\textbf{90.81} &\textbf{83.23} &\textbf{5.61} &\textbf{1.50} \\
    \bottomrule
  \end{tabularx}
  \label{tab:result_LA_5}
\end{table*}

\begin{figure*}[h]
  \centering
  \includegraphics[width=1.0\linewidth]{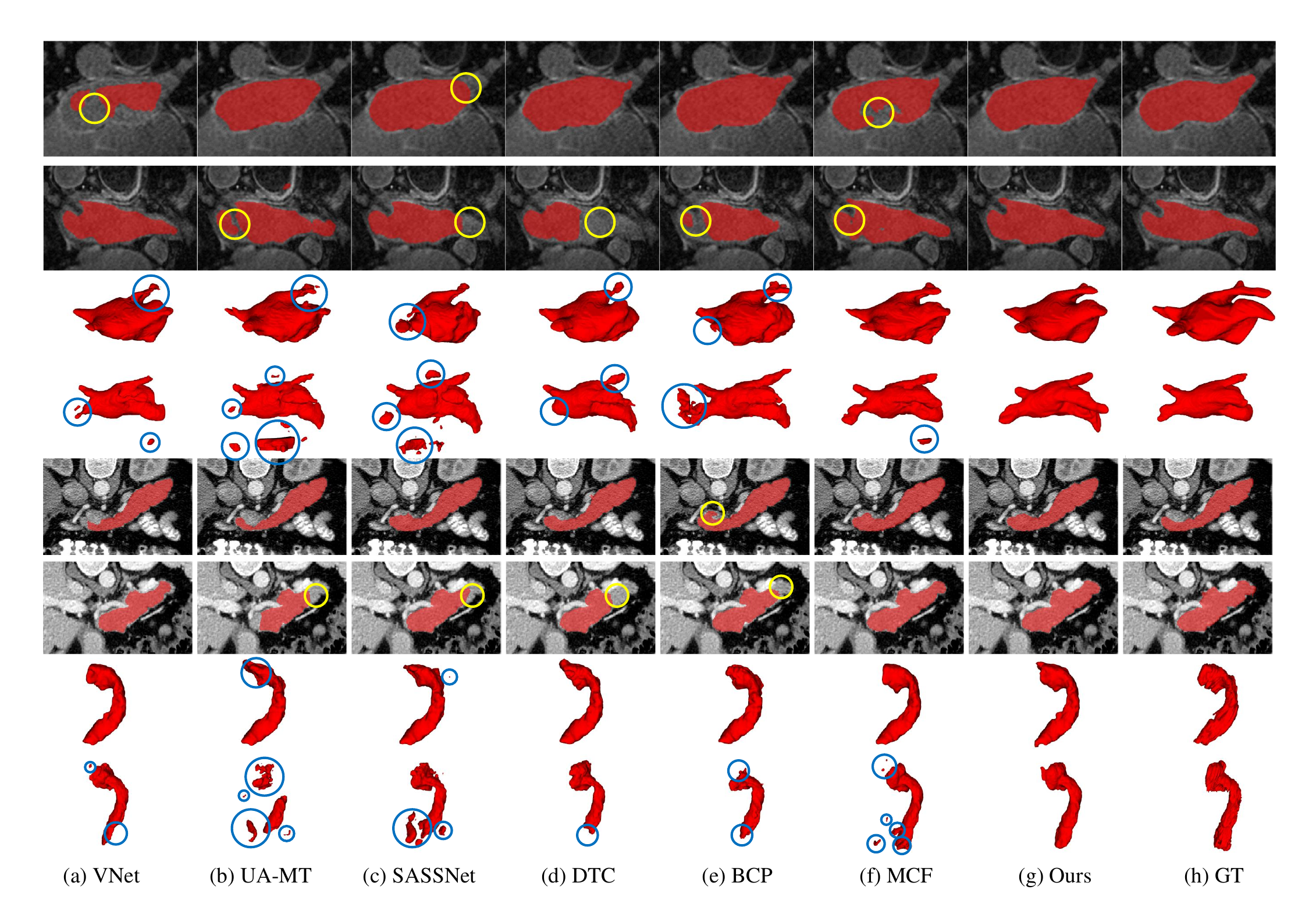}
  \caption{2D \& 3D segmentation visualization of different semi-supervised methods under 10\% labeled on LA (upper) and pancreas (bottom) dataset.}
  \label{fig:la2}
\end{figure*}

Our model outperformed previous models on all four metrics across 5\% and 10\% labeled data. At 5\% of the data, compared to the best results from previous work, our PMT showed a 1.45\% improvement in Dice, a 2.32\% improvement in Jaccard, a reduction of 1.45 in 95HD, and a reduction of 0.29 in ASD. Our PMT also showed a 1.19\% improvement in Dice, a 1.92\% improvement in Jaccard, a reduction of 1.20 in 95HD, and a reduction of 0.26 in ASD at 10\% of the data. It is noteworthy that our model maintains good performance even with a small amount of data. In comparison with state-of-the-art results, our model achieves a leading performance in the majority of metrics using only half of their data. For instance, we surpass MCF's performance at 10\% labeled data using only 5\% labeled data. The above indicates that our model achieves superior results compared to existing methods on differently proportioned annotated data in the task of segmenting the left atrium.

\textbf{Comparison on Pancreas-NIH Dataset.} We conducted a cross-model comparison on the classic Pancreas dataset. Detailed results of the experiments are presented in \textbf{Table~\ref{tab:result_pan_4}}. We tested with 10\% and 20\% of labeled data. To provide a more intuitive demonstration of the performance of various models on the Pancreas-NIH dataset, we have selected some representative results for visualization, as illustrated in \textbf{Fig.~\ref{fig:la2}}. Areas with inaccurate segmentation have been annotated accordingly. It is worth noting that the results in the table clearly indicate that Pancreas-NIH dataset is significantly more challenging than LA dataset. Therefore, we increased $\hat\lambda_{1}$ by a factor of two, resulting in improved performance.

\begin{table*}[h]
  \centering
  \setlength{\tabcolsep}{0pt}
  \caption{Comparison results on Pancreas-NIH dataset with 10\% and 20\% labeled data}
  \begin{tabularx}{0.92\textwidth}{@{}>{\centering\arraybackslash}p{0.25\textwidth}>{\centering\arraybackslash}p{0.15\textwidth}>{\centering\arraybackslash}p{0.13\textwidth}>{\centering\arraybackslash}p{0.13\textwidth}>{\centering\arraybackslash}p{0.13\textwidth}>{\centering\arraybackslash}p{0.13\textwidth}@{}}
    \toprule
    \multirow{2}{*}{Method} & \multirow{2}{*}{Labeled} & \multicolumn{4}{c}{Metrics} \\
     \cmidrule(lr){3-6}
    &  & Dice$\uparrow$ & Jaccard$\uparrow$ & 95HD$\downarrow$ & ASD$\downarrow$ \\
    \midrule
    VNet &6(10\%) &55.60 &41.74 &45.33 &18.63 \\
    UA-MT &6(10\%) &66.34 &53.21 &17.21 &4.57 \\
    SASSNet &6(10\%) &68.78 &53.86 &19.02 &6.26 \\
    DTC &6(10\%) &69.21 &54.06 &17.21 &5.95 \\
    $\text{BCP}^{\text{SOTA}}$ &6(10\%) &73.83 &59.24 &12.71 &3.72 \\
    $\text{MCF}^{\text{SOTA}}$ &6(10\%) &- &- &- &- \\
    PMT(Ours) &6(10\%) &\textbf{81.00} &\textbf{68.33} &\textbf{6.36} &\textbf{1.62} \\
    \midrule
    VNet &12(20\%) &72.38 &58.26 &19.35 &5.89 \\
    UA-MT &12(20\%) &76.10 &62.62 &10.84 &2.43 \\
    SASSNet &12(20\%) &77.66 &64.08 &10.93 &3.05 \\
    DTC &12(20\%) &78.27 &64.75 &8.36 &2.25 \\
    $\text{BCP}^{\text{SOTA}}$ &12(20\%) &82.91 &70.97 &\textbf{6.43} &2.25 \\
    $\text{MCF}^{\text{SOTA}}$ &12(20\%) &75.00 &61.27 &11.59 &3.27 \\
    PMT(Ours) &12(20\%) &\textbf{83.22} &\textbf{71.52} &{7.60} &\textbf{1.89} \\
    \bottomrule
  \end{tabularx}
  \label{tab:result_pan_4}
\end{table*}

Our model outperformed previous models on all four metrics across 10\% and 20\% labeled data. At 10\% of the data, compared to the best results from previous work, our PMT showed a 7.17\% improvement in Dice, a 9.09\% improvement in Jaccard, a reduction of 6.35 in 95HD, and a reduction of 2.10 in ASD. Our PMT also showed a 0.31\% improvement in Dice, a 0.55\% improvement in Jaccard, and a reduction of 0.36 in ASD at 20\% of the data. The above indicates that our model achieves superior results compared to existing methods on differently proportioned annotated data in the task of segmenting the pancreas.

\section{Conclusion}
\label{sec:Discussions}

\indent\indent In this paper, we propose a semi-supervised medical image segmentation framework named Progress Mean Teacher (PMT). PMT adopts a progressive design training process, establishing temporal-level model alignment and pseudo-label filtering while leveraging a network architecture based on MT. The core idea of this model framework is to generate diverse pseudo-labels consistently during the backpropagation process by establishing representation capability differences caused by iteration gaps, thereby stabilizing and enhancing the model's representation capability. In addition to the progressive architecture, PMT employs two simple yet effective methods, Pseudo Label Filtering~(PLF) and Discrepancy Driven Alignment~(DDA). PLF utilizes the representation capability of the model to filter pseudo-labels, discarding low-fidelity ones detrimental to co-training, while DDA aligns the differences in model predictions, allowing lagging models to catch up with leading models rapidly. Results from ablation experiments demonstrate that each component of the PMT framework significantly enhances the model's performance. In comparative experiments with other methods, PMT achieves state-of-the-art performance in terms of accuracy , surpassing previous methods significantly, and maintains this advantage compared to other methods in situations with more limited data and more challenging tasks.

\textbf{Limitation and Future Work.} Despite the outstanding performance of PMT, there is still room for further exploration of semi-supervised training architectures established through temporal consistency. Investigating more advanced and stable strategies under a progressive design paradigm, and assessing whether they can further enhance performance, are topics worthy of further research in the future.


\section*{Acknowledgments}

This work was supported in part by National Natural Science Foundation of China under Grants 62088102, 12326608 and 62106192, Natural Science Foundation of Shaanxi Province under Grant 2022JC-41, and Fundamental Research Funds for the Central Universities under Grant XTR042021005.

\bibliographystyle{splncs04}
\bibliography{main}
\end{document}